\documentclass{ifacconf}

\usepackage{graphicx}      
\usepackage{natbib}        

\usepackage{graphicx}%
\usepackage{multirow}%
\usepackage{amsmath,amssymb,amsfonts}%
\usepackage{mathrsfs}%
\usepackage[title]{appendix}%
\usepackage{xcolor}%
\usepackage{textcomp}%
\usepackage{manyfoot}%
\usepackage{booktabs}%
\usepackage{algorithm}%
\usepackage{algorithmicx}%
\usepackage{algpseudocode}%
\usepackage{listings}%
\usepackage{array}
\begin{document}
\begin{frontmatter}

\title{Robust Deterministic Policy Gradient for Disturbance Attenuation and Its Application to Quadrotor Control\thanksref{footnoteinfo}}

\thanks[footnoteinfo]{This work was supported by the Institute of Information Communications Technology Planning Evaluation (IITP) funded by the Korea government under Grant 2022-0-00469.}

\author[First]{Taeho Lee, Donghwan Lee} 
\address[First]{School of Electrical Engineering, Korea Advanced Institute of Science and Technology (KAIST), Daejeon 34141, South Korea (e-mail: eho0228@kaist.ac.kr, donghwan@kaist.ac.kr).}

\begin{abstract}                
This paper presents a robust reinforcement learning algorithm called robust deterministic policy gradient (RDPG), which reformulates the $H_\infty$ control problem as a two-player zero-sum dynamic game between a user and an adversary. The method combines deterministic policy gradients with deep reinforcement learning to train a robust policy that attenuates disturbances efficiently. A practical variant, robust deep deterministic policy gradient (RDDPG), integrates twin-delayed updates for stability and sample efficiency. Experiments on an unmanned aerial vehicle demonstrate superior robustness and tracking accuracy under severe disturbance conditions.
\end{abstract}


\end{frontmatter}

\section{Introduction}
In recent years, deep reinforcement learning (DRL) has shown remarkable success in finding optimal control policies across various domains, including games~\citep{RL1, DQN} and control tasks~\citep{DDPG, RL2}, where various model-free algorithms have demonstrated the ability to generate effective control inputs without requiring explicit system models~\citep{DDPG,TD3,SAC,PPO}. 
However, in real-world applications, physical systems are inevitably subject to external disturbances, unmodeled dynamics, and variations in system parameters. These factors often lead to significant discrepancies between the training simulation environments and the conditions encountered in practice~\citep{RARL,AR-MDP,HRL}. Consequently, policies trained in simulation frequently suffer from performance degradation or instability when deployed on real systems. This highlights that developing DRL policies that remain robust under unseen disturbances and model uncertainties is essential for ensuring reliability in practical applications.

In classical control theory, the $H_\infty$ control framework~\citep{H1,H2,Hinf1} provides a principled approach for designing controllers that minimize the worst-case impact of disturbances on system performance. A well-known result shows that, under certain assumptions, the $H_\infty$ control problem can be reformulated as a two-player zero-sum dynamic game (TZDG), or equivalently, a min–max optimization problem~\citep{Hbook}. In this game-theoretic formulation, the controller (first player) seeks to minimize a given cost function, while the adversary (second player) simultaneously tries to maximize it by injecting disturbances~\citep{Hbook}.

Building upon this theoretical foundation, we introduce the robust deterministic policy gradient (RDPG) algorithm, which enhances the robustness of the deterministic policy gradient (DPG) method~\citep{DPG} by transforming the $H_\infty$ control formulation into a TZDG between the user (controller) and an adversarial agent. We develop a reinforcement learning framework in which both agents are trained simultaneously: the user aims to minimize the objective, while the adversary attempts to maximize it. This min–max interplay naturally encourages the user to acquire a policy that is robust to worst-case disturbances.
Furthermore, we extend this idea to high-dimensional continuous control settings by incorporating it into a deep reinforcement learning framework. The resulting algorithm, named robust deep deterministic policy gradient (RDDPG), combines the core idea of RDPG with TD3~\citep{TD3}.

Additionally, we aim to develop a robust control strategy for quadrotors, a widely used unmanned aerial vehicle (UAV) in various applications. The ability to maintain robust control of quadrotors is particularly important in dynamic and uncertain environments, where tasks such as tracking moving targets or following predefined waypoints are required. To this end, we demonstrate that the proposed RDDPG method is capable of generating optimal control inputs even in the presence of external disturbances. Experimental results show that RDDPG achieves lower costs and exhibits stronger robustness under disturbance conditions compared to existing DRL-based control approaches.

In conclusion, the main contributions can be summarized as follows:
\begin{enumerate}
    \item We propose the RDPG algorithm, which enhances robustness against external disturbances by integrating concepts from $H_\infty$ control and two-player zero-sum dynamic games into the actor-critic framework.
    \item We develop a novel deep reinforcement learning algorithm, named RDDPG, which integrates RDPG into the TD3 framework to enable scalable and robust learning in continuous control tasks.
    \item We apply the proposed RDDPG algorithm to the tracking control of quadrotor and conduct comprehensive numerical simulations. The results demonstrate that RDDPG significantly outperforms several state-of-the-art DRL and robust RL methods.
\end{enumerate}

\section{Related works}\label{sec:related works}

The $H_\infty$ control methods are effective in designing robust controllers against disturbances and have demonstrated satisfactory performance in nonlinear dynamic systems~\citep{Hinf1, Hinf2, Hinf3,Hinf5}. However, these methods require precise model linearization and incur high computational costs due to the need to solve Hamilton-Jacobi-Isaacs (HJI) equations. To address these limitations, alternative approaches leverage deep reinforcement learning (DRL) to determine optimal control policies without the need for solving nonlinear HJI equations.

Recent efforts have extended the ideas of $H_\infty$ control and two-player zero-sum games into the DRL framework to enhance robustness under various scenarios. For instance, \citet{RRLforquad} developed a robust DRL approach to bridge the gap between different environments for quadcopter control tasks. Their method primarily addresses model uncertainties rather than external disturbances by adopting an action-robust reinforcement learning framework~\citep{AR-MDP}, where an adversary injects noise directly into the agent's actions to minimize the agent's reward. 
In contrast, our method explicitly introduces external disturbances into the environment dynamics and formulates the learning process as a two-player zero-sum game between the agent and the adversary. By doing so, our approach enables the agent to learn a control policy that is robust to worst-case disturbance scenarios, rather than merely compensating for action-level noise.
~\citet{HRL} proposed a novel approach that models robust locomotion learning as an interaction between the locomotion policy and a learnable disturbance generator. The disturbance is conditioned on the robot’s state and produces appropriate external forces to facilitate policy learning.
To address modeling errors and discrepancies between training and testing conditions, robust adversarial reinforcement learning (RARL)~\citep{RARL} was proposed. RARL trains the agent to perform effectively in the presence of a destabilizing adversary that applies disturbance forces to the system. The agent and the adversary are concurrently trained using the trust region policy optimization (TRPO) algorithm, with the agent learning to maximize the expected reward while the adversary learns to minimize it.
Although the underlying concept is similar, their approach differs significantly from ours. Both methods adopt stochastic policy frameworks and they do not impose explicit constraints on the adversary, which may lead to the generation of excessively strong or unrealistic disturbances during training. In contrast, our method is based on deterministic policy gradients and explicitly incorporates the $H_\infty$ control perspective into a two-player zero-sum game formulation, enabling more stable and practical learning in continuous control tasks.

\section{Preliminaries}\label{sec:preliminaries}

\subsection{$H_\infty$ Control} 

Let us consider the discrete time nonlinear discrete-time system
\begin{align*}
    x_{ k+1} &= f(x_k,u_k,w_k,v_k)\\
    y_k& = g(x_k,u_k,w_k)
\end{align*}
where $x_k \in \mathbb{R}^p$ is the state, $u_k\in \mathbb{R}^m$ is the control input, $w_k\in \mathbb{R}^n$ is the disturbance, and $v_k\in \mathbb{R}^l$ is the process noise at time step $k$.
Using the state-feedback controller $u=\pi(x)$, the system can be reduced to the autonomous closed-loop system
\begin{align*}
    x_{k+1} &= f(x_k,\pi(x_k),w_k,v_k)\\
    y_k &= g(x_k,\pi(x_k),w_k)
\end{align*}

Assume that the initial state $x_0$ is determined by $x_0 \sim \rho ( \cdot )$, where $\rho$ is the initial state distribution. 
Defining the stochastic processes ${\bf w}_{0:\infty}:= (w_0,w_1,\ldots)$ and ${\bf y}_{0:\infty}:= (y_0,y_1,\ldots)$, the system can be seen as a stochastic mapping from ${\bf w}_{0:\infty}$ to ${\bf y}_{0:\infty}$ as follows: ${\bf y}_{0:\infty} \sim T_{\pi}(\cdot|{\bf w}_{0:\infty})$, where $T_{\pi}$ is the conditional probability of ${\bf y}_{0:\infty}$ given ${\bf w}_{0:\infty}$.
Moreover, defining the $L^2$ norm for the general stochastic process ${\bf z}_{0:\infty}:= (z_0,z_1,\ldots)$ by 
\begin{align*}
||{{\bf{z}}_{0:\infty }}|{|_{{L^2}}}: = \sqrt {\sum\limits_{k = 0}^\infty  {\mathbb E}[ \|z_k \|_2^2]}, 
\end{align*}
The $H_\infty$ norm of the autonomous system is defined as
\begin{align}
||{T_\pi }|{|_\infty }: = \mathop {\sup }\limits_{{{\bf{w}}_{0:\infty }} \ne 0} \frac{{{{\left\| {{{\bf{y}}_{0:\infty }}} \right\|}_{{L^2}}}}}{{{{\left\| {{{\bf{w}}_{0:\infty }}} \right\|}_{{L^2}}}}}
\label{eq:Hnorm}
\end{align}
The goal of $H_\infty$ control is to design a control policy $\pi$ that minimizes the $H_\infty$ norm of the system $T_\pi$. However, minimizing $ ||T_\pi||_\infty$ directly is often difficult in practical implementations. Instead, we typically aim to find an acceptable $\eta > 0$ and a controller satisfying $ ||T_\pi||_\infty \leq \eta$, which is called suboptimal $H_\infty$ control problem. Then, the $H_\infty$ control problem can be approximated to the problem of finding a controller $\pi$ that satisfies the constraint
\begin{equation}
||{T_\pi }||_\infty ^2 = \mathop {\sup }\limits_{{{\bf{w}}_{0:\infty }} \ne 0} \frac{{||{{\bf{y}}_{0:\infty }}||_{{L^2}}^2}}{{||{{\bf{w}}_{0:\infty }}||_{{L^2}}^2}} \le {\eta ^2}
\end{equation}
Defining 
\begin{align}
{J^\pi }: = \mathop {\sup }\limits_{{\bf w}_{0:\infty} \ne 0}{\mathbb  E}\left[ {\left. {\sum\limits_{k = 0}^\infty  {( \|y_k \|_2^2  - \eta ^2 \|w_k \|_2^2 )} } \right|\pi} \right]
\label{eq:H_J}
\end{align}
the problem can be equivalently written by finding a controller $\pi$ satisfying ${J^\pi } \le 0$. The problem can be solved by 
\begin{align*}
{\pi ^*}: = \arg \mathop {\min }\limits_\pi  {J^\pi }.
\end{align*}
 
\subsection{Two-player zero-sum dynamic game}
According to \citep{Hbook}, the suboptimal $H_\infty$ control problem can be equivalently viewed as solving a two-player zero-sum dynamic game (TZDG) under certain assumptions. In particular, we consider two decision making agents called a user and an adversary, respectively. TZDG can be expressed as a tuple $(\mathcal{S},\mathcal{A}_1,\mathcal{A}_2,P,c,\gamma,x_0)$. Here $\mathcal{S}=\mathbb{R}^p$ is the state space, $\mathcal{A}_1=\mathbb{R}^m$ and $\mathcal{A}_2=\mathbb{R}^n$ are the continuous action spaces for the user and adversary, respectively. The user selects an control input $u \in \mathcal{A}_1$ and the second agent selects disturbance $ w \in \mathcal{A}_2$  at the current state $x \in \mathcal{S}$ simultaneously. Then the state transits to the next state $x' \in \mathcal{S}$ with the state transition probability $P(x'|x,u,w)$ and the cost $c\in \mathbb{R}$ is incurred by the cost function $c(x,u,w,x'): \mathcal{S} \times \mathcal{A}_1 \times \mathcal{A}_2 \times \mathcal{S} \rightarrow \mathbb{R}$. The $\gamma\in (0,1]$ is the discounted factor and $x_0\in \mathcal{S}$ represents the initial state.

We consider the state-feedback deterministic control policy for each agent, defined as $\pi:{\mathbb R}^p \to {\mathbb R}^m$ for the user and $\mu:{\mathbb R}^p \to {\mathbb R}^n$ for the adversary
\begin{align*}
    u = \pi (x) \in  {\mathbb R}^m,\quad w = \mu (x) \in {\mathbb R}^n,\quad x \in  {\mathbb R}^p
\end{align*}
The objective of the user and adversary are to minimize and maximize the cumulative discounted costs over infinite time horizon $J^{\pi ,\mu }$ defined as
\begin{align*}
    J^{\pi ,\mu }: = {\mathbb E}\left[ \left. \sum\limits_{k = 0}^\infty  \gamma^k c(x_k,u_k,w_k,x_{k+1}) \right|\pi,\mu\right]
\end{align*}
where $\mathbb{E}[\cdot|\pi,\mu]$ is an expectation conditioned on the two policies $\pi$ and $\mu$. 
The goal of the dynamic game is to find a saddle-point pair of equilibrium policies $(\pi,\mu)$ (if exists) for which 
\begin{align*}
    {J^{{\pi ^*},\mu }} \le {J^{\pi *,\mu *}} \le {J^{\pi ,\mu *}},\quad \forall \pi  \in \Pi ,\quad \forall \mu  \in {\rm M}
\end{align*}
where $\Pi$ denotes the set of all admissible state-feedback policies for the user and ${\rm M}$ denotes the set of all admissible state-feedback policies for the adversary. The above relation equivalently means 
\begin{align*}
    {\mu ^*} = {\max _{\mu  \in {\rm M}}}{J^{{\pi ^*},\mu }},\quad {\pi ^*} = {\min _{\pi  \in \Pi }}{J^{\pi ,{\mu ^*}}}
\end{align*}
\citet{perolat2015approximate,AR-MDP} demonstrated that for a game with optimal equilibrium return $J^*$, there always exists the Nash equilibrium, and it is equivalent to the minimax solution,
\begin{align}
    {J^*}: = {J^{{\pi ^*},{\mu ^*}}} = {\max _{\mu  \in {\rm M}}}{\min _{\pi  \in \Pi }}{J^{\pi ,\mu }} = {\min _{\pi  \in \Pi }}{\max _{\mu  \in {\rm M}}}{J^{\pi ,\mu }}
\end{align}

To show a connection between this TZDG framework and the $H_\infty$ control problem, let us consider the specific per-step cost function 
\begin{align*}
    c(x_k,u_k,w_k,x_{k+1}) = \left\| {g(x_k,u_k,w_k,x_{k+1})} \right\|_2^2 - {\eta ^2}\left\| w_k \right\|_2^2.
\end{align*}
Then the corresponding cost function is then written when $\gamma=1$ as 
\begin{align}
    &{J^{\pi ,\mu }} \nonumber = {\mathbb E}\left[ \left. \sum\limits_{k = 0}^\infty  \gamma^k c(x_k,u_k,w_k,x_{k+1}) \right|\pi,\mu\right]\\
    &= {\mathbb E}\left[ {\left. {\sum\limits_{k = 0}^\infty  {(  \|g(x_k,u_k,w_k,x_{k+1}) \|_2^2  - \eta ^2 \|w_k \|_2^2)} } \right|\pi ,\mu } \right].
\label{eq:TZDG_J}
\end{align}
The user's optimal policy is 
\begin{align*}
    {\pi ^*}: = \arg {\min _{\pi  \in \Pi }}{\max _{\mu  \in {\rm M}}}{J^{\pi ,\mu }}
\end{align*}
which is structurally very similar to Equation (\ref{eq:H_J}), the policy with the $H_\infty$ performance in the previous subsection. It is known that under some special conditions such as the linearity, the $H_\infty$ suboptimal control policy and the saddle-point policy are equivalent when $J^*\leq 0$. 

\subsection{Bellman equations}
Let us define the value function
\begin{align*}
    {V^{\pi ,\mu }}(x): = {\mathbb E}\left[ {\left. {\sum\limits_{k = t}^\infty \gamma^k c(x_k,u_k,w_k,x_{k+1}) } \right|\pi ,\mu ,x_k = x} \right]
\end{align*}
so that the corresponding cost function is $J^{\pi ,\mu} = {\mathbb E}[V^{\pi ,\mu }(x)|x \sim \rho ( \cdot )]$. We can also prove that the value function satisfies the Bellman equation 
\begin{align*}
    V^{\pi,\mu}(x) = c(x,\pi (x),\mu (x),x') +\gamma {\mathbb E}[{V^{\pi ,\mu }}(x')|\pi ,\mu ]
\end{align*}
where $x'$ implies the next state given the current state $x$ and the action taken by $(\pi ,\mu)$. Let us define the optimal value function as $V^*:=V^{\pi^*,\mu^*}$. Then, the corresponding optimal Bellman equation can be obtained by replacing $(\pi,\mu)$ by $(\pi^*,\mu^*)$. Similarly, let us define the so-called Q-function by
\begin{align*}
&{Q^{\pi ,\mu }}(x,u,w) := \\
&{\mathbb E}\left[ {\left. {\sum\limits_{k = t}^\infty  {\gamma^k c(x_k, u_k, w_k, x_{k+1})} } \right|{x_k} = x,{u_k} = u,{w_k} = w,\pi ,\mu } \right]
\end{align*}
which satisfies the Q-Bellman equation 
\begin{align*}
    {Q^{\pi ,\mu }}(x,u,w) = c(x,u,w,x') + \gamma {\mathbb E}[{Q^{\pi ,\mu }}(x',u',w')|\pi ,\mu ]
\end{align*}
where $x'$ implies the next state given the current state $x$ and action pair $(u,w)$, $u'$ means the next action of the user given $x'$ and under $\pi$, and $w'$ implies the next action of the adversary given $x'$ and under $\mu$. Defining the optimal Q-function $Q^*:=Q^{\pi^*,\mu^*}$, one can easily prove the optimal Q-Bellman equation
\begin{align*}
    {Q^*}(x,u,w) = c(x,u,w,x') +\gamma {\mathbb E}[{Q^*}(x',u',w')|{\pi ^*},{\mu ^*}]
\end{align*}

\section{Proposed method}\label{sec:method}

\begin{figure}[!t]
    \centering
    \includegraphics[width=1\linewidth]{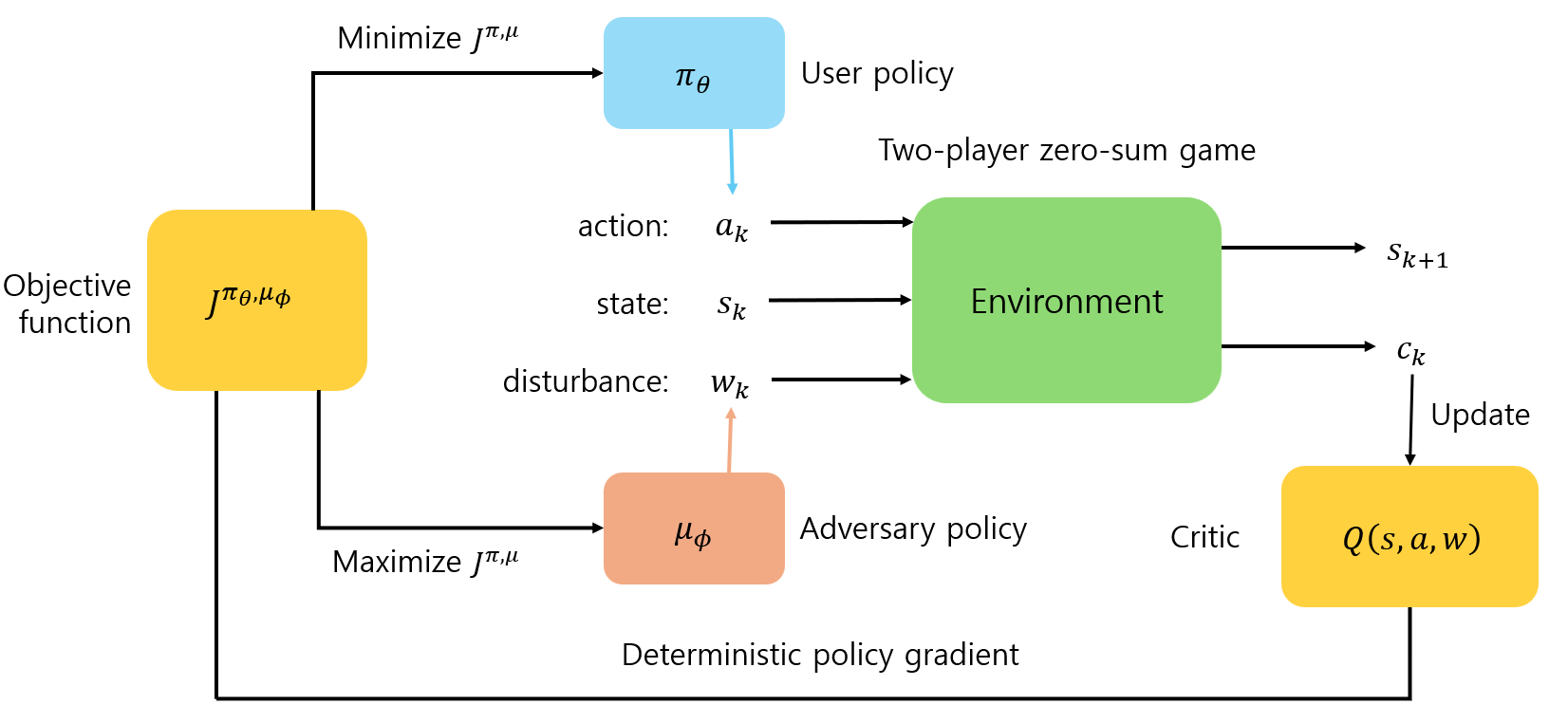}
    \caption{Overview of RDDPG. Based on the two-player zero-sum dynamic game, the user policy $\pi_\theta$ seeks to minimize the objective function $J^{\pi_\theta,\mu_\phi}$, while the adversary policy $\mu_\phi$ attempts to maximize it by injecting disturbances into the environment.}
    \label{fig:RDDPG}
\end{figure}

In this section, we will present the proposed method, robust deterministic policy gradient (RDPG) and its deep reinforcement learning version, robust deep deterministic policy gradient (RDDPG).
Fig.~\ref{fig:RDDPG} illustrates the architecture of the proposed method, RDDPG, based on the TZDG. In this method, two policy, user policy $\pi_\theta$ and adversary policy $\mu_\phi$, interact within the environment. The user policy generates the action $u_k$, while the adversary policy generates the disturbance $w_k$, both based on the current state $x_k$. These inputs are fed into the environment, which returns the next state $x_{k+1}$ and the cost $c_{k+1}$ based on the dynamics of the environment. The critic is updated based on ${x_k,u_k,w_k,x_{k+1},c_{k+1}}$ and evaluates the joint performance of the user and adversary. The policy of user and adversary are updated by deterministic policy gradient method to minimize and maximize the objective function $J^{\pi_\theta,\mu_\phi}$. The detailed procedure is described in the following sections.

\subsection{Robust deterministic policy gradient}
To apply deterministic policy gradient (DPG)~\citep{DPG} to the two-player zero-sum dynamic games, we consider the parameterized deterministic control policies 
\[u = \pi_{\theta} (x) \in {\mathbb R}^m,\quad w = \mu_{\phi} (x) \in {\mathbb R}^n,\quad x \in {\mathbb{R}^p}\]
where $\pi_\theta: \mathbb{R}^p \rightarrow \mathbb{R}^m$ denotes the user’s control policy parameterized by $\theta$ and $\mu_\phi: \mathbb{R}^p \rightarrow \mathbb{R}^n$ denotes the adversary’s control policy parameterized by $\phi$.
The saddle-point problem is then converted to 
\[{\theta ^*}: = \arg {\min _\theta }{\max _\phi }{J^{{\pi _\theta },{\mu _\phi }}}\]
which can be solved using the primal-dual iteration
\begin{align*}
{\theta _{k + 1}} =& {\theta _k} - {\alpha _\theta }{\left. {{\nabla _\theta }{J^{{\pi _\theta },{\mu _{{\phi _k}}}}}} \right|_{\theta  = {\theta _k}}},\\
{\phi _{k + 1}} =& {\phi _k} + \alpha_\phi{\left. {{\nabla _\phi }{J^{{\pi _{{\theta _{k + 1}}}},{\mu _\phi }}}} \right|_{\phi  = {\phi _k}}}
\end{align*}
where $\alpha_\theta>0$ and $\alpha_\phi>0$ are step-sizes.
According to~\citet{DPG}, the deterministic gradients can be obtained using the following theorem.
\begin{thm}
    The deterministic policy gradients for the user and adversary policies are given by
\[{\nabla _\theta }{J^{{\pi _\theta },{\mu _\phi }}} = {\mathbb E}\left[ {\left. {{{\left. {{\nabla _\theta }{Q^{\pi ,{\mu _\phi }}}(s,{\pi _\theta },{\mu _\phi })} \right|}_{\pi  = {\pi _\theta }}}} \right|s \sim \rho} \right]\]
and
\[{\nabla _\phi }{J^{{\pi _\theta },{\mu _\phi }}} = {\mathbb E} \left[ {\left. {{{\left. {{\nabla _\phi }{Q^{{\pi _\theta },\mu }}(s,{\pi _\theta },{\mu _\phi })} \right|}_{\mu  = {\mu _\phi }}}} \right|s \sim \rho} \right]\]
respectively.
\end{thm}
The proof is a simple extension of the DPG theorem~\citep{DPG}, so it is omitted here. A reinforcement learning counterpart to DPG can be obtained by using the samples of the gradients. Next, we will consider the following per-step cost function:
\begin{align*}
    c(x_k,u_k,w_k,x_{k+1}) = \tilde c(x_k,u_k,w_k,x_{k+1}) - \eta ^2 \left\| w_k \right\|_2^2
\end{align*}
where $\tilde c:{\mathbb R}^p\times {\mathbb R}^m\times {\mathbb R}^n\times {\mathbb R}^p\to {\mathbb R}$ is the cost function for the user. Note that the cost function $\tilde c$ can be set arbitrarily and is more general than the quadratic cost in the $H_\infty$ control problem. This is because the $H_\infty$ control problem can be recovered as shown Equation (\ref{eq:H_J}) and (\ref{eq:TZDG_J}) by setting 
\begin{align*}
    \tilde c(x_k,u_k,w_k,x_{k+1}) = \left\| {g(x_k,u_k,w_k,x_{k+1})} \right\|_2^2
\end{align*}

In this way, we can consider the TZDG as an extension of the usual reinforcement learning tasks by considering robustness against the adversarial disturbance. The additional disturbance quadratic cost ${\eta^2}w^T w$ in the cost function discourages the adversary from generating excessively large disturbances, effectively regularizing its behavior and ensuring that the learned policies are robust against realistic worst-case scenarios rather than arbitrary or unbounded inputs.

Let us assume that we found some approximate solutions $\theta^*_\epsilon,\phi^*_\epsilon$ such that $J^{\pi _{\theta _\varepsilon ^*},\mu _{\phi _\varepsilon ^*}} \le 0$. Then, this implies that 
\[\frac{{{\mathbb E}\left[ {\left. {\sum\limits_{k = 0}^\infty  {\gamma ^k \tilde c({x_k},{u_k},{w_k},{x_{k+1}})} } \right|{\pi _{\theta _\varepsilon ^*}},{\mu _{\phi _\varepsilon ^*}}} \right]}}{{{\mathbb E} \left[ {\left. {\sum\limits_{k = 0}^\infty  \gamma ^k \left\| w_k \right\|_2^2} \right|{\pi _{\theta _\varepsilon ^*}},{\mu _{\phi _\varepsilon ^*}}} \right]}} \le {\eta ^2}\]

\subsection{Robust deep deterministic policy gradient }

To implement RDPG in high-dimensional continuous control tasks, we incorporate the techniques from twin-delayed deep deterministic policy gradient (TD3) \citep{TD3}, resulting in the RDDPG. In particular, we introduce the following networks:
\begin{enumerate}
    \item The online actor networks $\pi_{\theta}(x)$ and $\mu_\phi(x)$ for the user adversary respectively
    \item The corresponding target actor networks $\pi_{\theta'}(x)$, $\mu_{\phi'}(x)$
    \item Two online critic networks $Q_{\psi_1}(x,u,w) , Q_{\psi_2}(x,u,w)$
    \item The corresponding target critic networks 
    
    $Q_{\psi_1'}(x,u,w)$, $Q_{\psi_2'}(x,u,w)$ 
\end{enumerate}
Now, following~\cite{TD3}, the actor and critic networks are trained through the following procedures.

\subsubsection{Critic update}
The critic is trained by the gradient descent step to the loss
\begin{align*}
L_{\rm critic}&(\psi_i;B)  :=\\ & \frac{1}{|B|}\sum\limits_{(x,u,w,c,x') \in B} {(y - Q_{\psi_i}(x,u,w))}^2,
i \in \{ 1,2\}
\end{align*}
where $B$ is the mini-batch, $|B|$ is the size of the mini-batch. Moreover, the target $y$ is defined as
\[y = c + \gamma \min_{i \in \{ 1,2\} }Q_{\psi_i'}(x',\tilde u,\tilde w)\]
when $x'$ is not the terminal state, and 
\[y = c \]
when $x'$ is the terminal state (in the episodic environments), where 
\begin{align*}
    \tilde{u} &= \pi_{\theta'}(x') + {\rm clip}({\epsilon _1}, - \delta ,\delta) \ \ \epsilon_1 \sim \mathcal{N}(0,\sigma) \\
    \tilde{w}  &= \mu_{\phi'}(x') +{\rm clip}({\epsilon_2}, - \delta,\delta) \ , \ \epsilon_2 \sim \mathcal{N}(0,\sigma)
\end{align*}
where $\epsilon_1,\epsilon_2$ are random noise vectors added in order to smooth out the target values, $\mathcal{N}(0,\sigma)$ implies the Gaussian distribution with zero mean and variance $\sigma$,  and the clip function is added in order to guarantee the boundedness of the control inputs with bound $\delta$. The critic's online parameters $\psi_1$ and $\psi_2$ are updated by the gradient descent step to minimize the loss $L_{\rm critic}(\psi_i;B)$
\[{\psi _i} \leftarrow {\psi _i} - {\alpha _{{\rm{critic}}}}{\nabla _{\psi_i} }{L_{\rm critic}}({\psi _i};B),\quad i \in \{ 1,2\} \]

\subsubsection{Actor update}
The actor networks for the user and adversary are updated using the sampled deterministic policy gradient
\begin{align*}
\theta  \leftarrow \theta  - {\alpha _{\rm actor}}{\nabla _\theta }{L_{\rm actor}}(\theta ,\phi ;B)\\
\phi  \leftarrow \phi  + {\beta_{\rm actor}}{\nabla _\phi }{L_{\rm actor}}(\theta ,\phi ;B)
\end{align*}
where $\alpha_{\rm actor}$ and $\beta _{\rm actor}$ are the step-sizes, and  
\[L_{\rm actor}(\theta,\phi;B) = \frac{1}{{|B|}}\sum\limits_{(x,u,w,r,x') \in B} {{Q_{{\psi _1}}}(x,{\pi _\theta }(x),{\mu _\phi }(x))} \]
Please note the opposite signs of the gradients for the user and adversary updates, which are due to their opposite objectives. 


\begin{algorithm}[tb]
\caption{RDDPG}\label{alg:algorithm1}
\begin{algorithmic}[1]
\State Initialize the online critic networks $Q_{\psi_1},Q_{\psi_2}$
\State Initialize the actor networks $\pi_{\theta},  \mu_{\phi}$ for the user and adversary, respectively.
\State Initialize the target parameters $\psi_1'\gets\psi_1, \psi_2' \gets \psi_2$, $\theta' \gets \theta , \phi' \gets \phi$
\State Initialize the replay buffer $\mathcal{D}$

\For{ Episode $i$=1,2,...$N_{iter}$}
\State Observe $s_0$
\For{ Time step $k$=0,1,2,...$\tau$-1}
\State Select actions $u_k = \pi_{\theta}(x_k)+e_1$ and $w_k = \mu_{\phi}(x_k)+e_2$, \\ \quad\quad\quad where $e_1,e_2 \sim \mathcal{N}(0,\sigma)$ are exploration noises.
\State Compute the cost $c_{k+1}:=c(x_k,u_k,w_k)$
\State Observe the next state $x_{k+1}$

\State Store the transition tuple $(x_k,u_k,w_k,c_{k+1},x_{k+1})$ in the replay buffer $D$
\State Uniformly sample a mini-batch $B$ from the replay buffer $\mathcal{D}$
\State Update critic network:
\[{\psi _i} \leftarrow {\psi _i} - {\alpha _{{\rm{critic}}}}{L_{\rm critic}}({\psi _i};B),\quad i \in \{ 1,2\} \]
\State Update actor networks by the deterministic policy gradient:
\begin{align*}
\theta  \leftarrow \theta  - {\alpha _{\rm actor}}{\nabla _\theta }{L_{\rm actor}}(\theta ,\phi ;B)\\
\phi  \leftarrow \phi  + {\beta_{\rm actor}}{\nabla _\phi }{L_{\rm actor}}(\theta ,\phi ;B)    
\end{align*}

\State Soft update target networks:
\begin{align*}
    \theta' \gets \tau\theta +(1-\tau)\theta' \\
    \phi_i' \gets \tau\phi +(1-\tau)\phi'
\end{align*}
\EndFor
\EndFor
\end{algorithmic}
\end{algorithm}

\section{Quadrotor application}
\subsection{State space}
The state $x_k$ of the quadrotor is defined as
\begin{align*}
x_k = [\mathbf{e}_k^{p},\mathbf{e}_k^{v},\boldsymbol{q}_k,\boldsymbol{\omega}_k]^T \in \mathbb{R}^{13}
\end{align*}
where $\mathbf{e}_k^{p}$ denotes the position error, $\mathbf{e}_k^{v}$ represents the velocity error, $\boldsymbol{q}_k$ is the quaternion vector, and $\boldsymbol{\omega}_k$ is the angular velocity vector. These quantities are defined as follows:
\begin{align*}
    \mathbf{e}_k^{p} &= \mathbf{p}_k^{\text{target}} - \mathbf{p}_{k} \\
    & = [p_{x,k}^\text{target}-p_{x,k}, \ p_{y,k}^\text{target}-p_{y,k}, \ p_{z,k}^\text{target}-p_{z,k}]^T \in \mathbb{R}^3\\
    \mathbf{e}_k^{v} &= \mathbf{v}_k^{\text{target}} - \mathbf{v}_k \\ 
    & = [v_{x,k}^\text{target} - v_{x,k}, \ v_{k,y}^\text{target} - v_{y,k}, \ v_{z,k}^\text{target} - v_{z,k}]^T \in \mathbb{R}^3\\
    \boldsymbol{q}_k &= [q_{w,k},q_{x,k},q_{y,k}q_{z,k}]^T \in \mathbb{R}^4 \\
    \boldsymbol{\omega}_k &= [\omega_{\phi,k},\omega_{\theta,k},\omega_{\psi,k}]^T \in \mathbb{R}^3
\end{align*}
where $\mathbf{p}_k^{\text{target}}$ and $\mathbf{p}_k$ denote the position vectors of the target and the quadrotor, respectively, and $\mathbf{v}_k^{\text{target}}$ and $\mathbf{v}_k$ represent their corresponding linear velocity vectors.
As in the method used in deep Q-network (DQN)~\citep{DQN}, we used the last four time steps of the state as input to the neural networks to help the network better capture dynamics in the environment.

\subsection{Action space}

The action of the quadrotor, $u_k$, is the speed of four motors $\text{RPM}_{i,k}, \ i\in \{1,2,3,4\}$, which is continuous value between $0$ and $21713.714$, as follows:
\begin{align*}
    u_k :=[\text{RPM}_{1,k},\text{RPM}_{2,k},\text{RPM}_{3,k},\text{RPM}_{4,k}]^T  \in \mathbb{R}^4. 
\end{align*}
The disturbance of the adversary, $w_k$, represents the external forces to the quadrotor along the $x$, $y$, and $z$ axes of the body frame of the quadrotor as follows:
\begin{align*}
w_k := [w_{x,k},w_{y,k},w_{z,k}]^T \in \mathbb{R}^3
\end{align*}
For training stability, each component of the disturbance was constrained to lie within the range $[-0.1,,0.1]$.

\subsection{Cost function}
The cost function $c(x_k,u_k,w_k,x_{k+1})$ is comprised of the cost function for the user $\tilde{c}(x_k,u_k,w_k,x_{k+1})$ and the disturbance quadratic cost $\eta^2w_k^Tw_k$. The cost function for the user $\tilde{c}(x_k,u_k,w_k,x_{k+1})$ is the sum of the following terms:
$\tilde{c}(x_k,u_k,w_k,x_{k+1}) = c_{p,k} + c_{v,k} + c_{\rho,k} + c_{tr}$ 

\begin{enumerate}
    \item Position error cost $c_{p,k}$ : Position error between the target and quadrotor 
    \begin{equation*}
        c_{p,k} = \alpha \times ||\mathbf{e}_k^{p}||_2^2
    \end{equation*}
    \item Velocity error cost $c_{v,k}$ : Velocity error between the target and quadrotor
    \begin{equation*}
        c_{v,k} = \beta \times || \mathbf{e}_k^v ||_2^2
    \end{equation*}
    \item Angle cost $c_{\rho,k}$ : Magnitude of roll $\phi_k$, pitch $\theta_k$, and yaw $\psi_k$ of the quadrotor 
    \begin{equation*}
        c_{\rho,k} = \epsilon \times || \boldsymbol{\rho}_k ||_2^2
    \end{equation*}
    \item Penalty cost $c_{tr}$ : Penalty when the quadrotor fails to track the target
    \begin{equation*}
       c_{tr} = 
       \begin{cases}
    \zeta & \text{if } ||\mathbf{e}_k^{p}||_2 \geq 5 \text{  or } \phi_k, \theta_k \geq \frac{\pi}{2}\\
      0 & \text{else}
    \end{cases}
    \end{equation*}
\end{enumerate}
where the scaling factors $\alpha$, $\beta$, $\epsilon$, and $\zeta$ are positive (i.e., $\alpha, \beta, \epsilon, \zeta > 0$), and $\lambda$ is negative (i.e., $\lambda < 0$). The specific values used for these parameters are summarized in the Appendix of arxiv version.

\section{Experiments and results}
\label{sec:results}
\subsection{Experiments setups}
\begin{figure*}[!ht]
    \centering
    \includegraphics[width=1.0\linewidth]{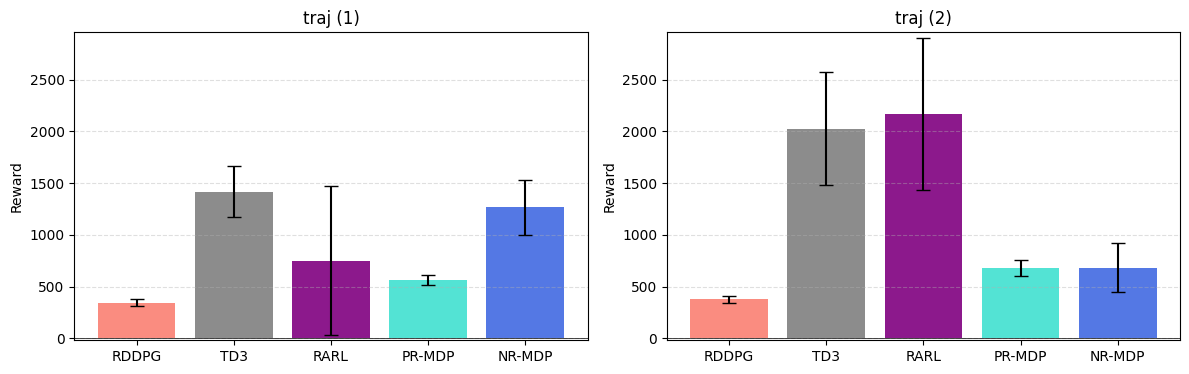}
    \caption{Mean and standard deviation of the total episode cost. RDDPG achieves the lowest average cost and the smallest variance. This indicates that the RDDPG yields a robust agent that performs consistently even under external disturbances.}
    \label{fig:cost}
\end{figure*}

\begin{figure}[t]
    \centering
    \begin{minipage}{0.24\textwidth}
        \centering
        \includegraphics[width=\textwidth]{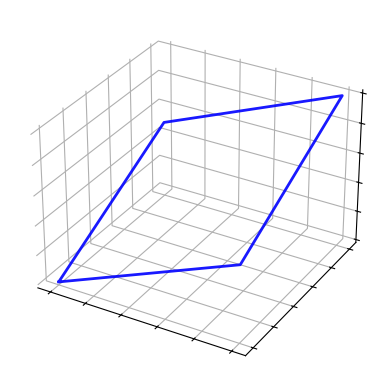}
        {\fontsize{8pt}{9pt} \selectfont \text{Traj (1)}}
    \end{minipage}
    \begin{minipage}{0.24\textwidth}
        \centering
        \includegraphics[width=\textwidth]{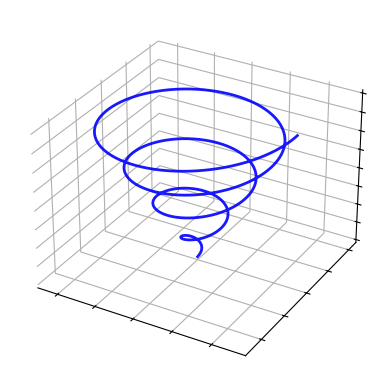}
        {\fontsize{8pt}{9pt} \selectfont \text{Traj (2)}}
    \end{minipage}
    \caption[Evaluation trajectories]{Two evaluation trajectories, denoted as Traj (1) and Traj (2), shown from left to right. } \label{fig:trajectories}
\end{figure}

For experiments, we used the gym-pybullet-drones environment~\citep{gym}, which is an open-source quadrotor simulator built with Python and the Bullet Physics engine. This environment provided a modular and precise physics implementation, supporting both low-level and high-level control. In the simulator, the Crazyflie 2.0 quadrotor was modeled, and its parameters were summarized in the accompanying arXiv version of the paper.

To evaluate the robustness of the proposed RDDPG algorithm, we compare it against deterministic policy gradient–based reinforcement learning methods, including TD3~\citep{TD3}, which employs a single agent that seeks to minimize the cost function~$\tilde{c}$. Beyond these standard single-agent baseline, we further benchmark RDDPG against robust reinforcement learning algorithms such as RARL~\citep{RARL}, PR-MDP, and NR-MDP~\citep{AR-MDP}. These robust methods are formulated under the TZDG framework, where the user and adversary aim to minimize and maximize the cost function, respectively.
The full set of hyperparameters used in our experiments is provided in the accompanying arXiv version of the paper.

For evaluation, we generated two trajectories as shown in Fig.~\ref{fig:trajectories}. 
To assess robustness to external perturbations, we conducted experiments in which the quadrotor was required to track these trajectories while being subjected to the disturbances $w_x,w_y,w_z$ were randomly sampled from the interval $[-0.2,0.2]$.
Each algorithm was evaluated over 500 episodes and assessed using the mean and standard deviation of the total cost per episode, where the total cost was computed as the sum of the per-step costs $\tilde{c}(x_k, u_k,w_k,x_{k+1})$ over a single episode.

\subsection{Results}

Figure~\ref{fig:cost} compares the performance of RDDPG with TD3, RARL~\citep{RARL}, PR-MDP, and NR-MDP ~\citep{AR-MDP} across two evaluation trajectories under random disturbances. Across both trajectories experiments, RDDPG consistently demonstrates the most stable and reliable performance. Its cost distribution exhibits the smallest variance among all evaluated methods, indicating that the learned policy maintains robust tracking behavior even under random disturbances. In contrast, TD3 and RARL often achieve higher peak rewards but suffer from significantly larger variance, revealing high sensitivity to perturbations and unstable control behaviors. PR-MDP and NR-MDP show more moderate performance with reduced variability compared to TD3 and RARL, yet they still fall short of the consistency exhibited by RDDPG.

By maintaining a consistently low and stable cost across a wide range of disturbance intensities, RDDPG exhibits superior robustness and reliable control performance, even under challenging environmental conditions. This enhanced robustness stems from its adversarial training framework, inspired by $H_\infty$ control theory and formulated as a two-player zero-sum game. By optimizing the policy against worst-case disturbances during training, RDDPG learns disturbance-resilient strategies that improve generalization and ensure stable performance under unseen or highly uncertain conditions.

\section{Conclusion}
\label{sec:conclusion}

This paper proposes RDPG, which combines the concept of the $H_\infty$ control problem and a two-player zero-sum dynamic game framework, to overcome the robustness problem of DRL algorithms. In RDPG, the user seeks to minimize the cost function while the adversary attempts to maximize it, thereby enabling the agent to learn a control policy that is robust to external disturbances. Furthermore, we introduce RDDPG, which integrates the robustness framework of RDPG with the learning stability of TD3, achieving both improved disturbance resistance and efficient policy learning in continuous control tasks.
In order to evaluate the robustness of RDDPG, we implement it on quadrotor tracking tasks under various external disturbances. Experimental results show that RDDPG successfully learns an optimal control policy while maintaining stable performance across different disturbance scenarios. These findings demonstrate that RDDPG outperforms other DRL-based methods in terms of robustness and reliability.

\bibliography{ifacconf}             
                                                   
\clearpage
\appendix
\section{Hyperparameters Used in Training}
\subsection{Physical parameters of Crazyflies 2.0}
\begin{table}[h]
\caption[Physical parameters of the Crazyfile 2.0]{Physical parameters of the Crazyfile 2.0}
\label{tab:crazyfiles}
\begin{tabular} {|c|c|c|}
\hline
Parameter & Description & Value \\
\hline
m & Total mass & $0.027$[kg] \\
\hline
l & Arm length & $0.0397$[m]\\
\hline
$K_F$ & Thrust coefficient & $3.16\times 10^{-10}$\\
\hline
$K_M$ & Torque coefficient & $7.94 \times 10^{-12}$\\
\hline
$I_{xx}$ & Principal Moment of Inertia around x axis & $1.4 \times 10^{-3}$ [$\text{kg} \times \text{m}^2$] \\
\hline
$I_{yy}$ & Principal Moment of Inertia around y axis & $1.4 \times 10^{-3}$ [$\text{kg} \times \text{m}^2$] \\
\hline
$I_{zz}$ & Principal Moment of Inertia around z axis & $2.17 \times 10^{-3}$ [$\text{kg} \times \text{m}^2$] \\
\hline
$K_{D.xy}$ & Drag coefficients in $x,y$ axes  & $9.18 \times 10^{-7}$ \\
\hline
$K_{D.z}$ &  Drag coefficients in $z$ axis & $10.31 \times 10^{-7}$ \\
\hline
\end{tabular}
\end{table}

\subsection{Cost function}
\begin{table}[!h]
\centering
\caption{Scaling factors in cost function}
\label{tab:cost_parameters}    
    \begin{tabular}{|c|c|c|}
    \hline
    Parameter & Description &Value \\
    \hline
    $\alpha$& Coefficient for position error & $10$ \\
    \hline
    $\beta$ & Coefficient for velocity error & $1$ \\
    \hline
    $\epsilon$ & Coefficient for angle error & $1$ \\
    \hline
    $\zeta$ & Penalty & $1000$ \\
    \hline
\end{tabular}
\end{table}

\subsection{Algorithm parameters}
\begin{table}[h!]
    \caption{Algorithm parameters}
    \centering
    \begin{tabular}{|c|c|c|}
    \hline
    Parameter & Description & Value \\
    \hline
        $\alpha_{actor}$ & Learning rate for user& 0.0001 \\
        \hline
        $\beta_{actor}$ & Learning rate for adversary& 0.0001 \\
        \hline
        $\alpha_{critic}$ & Learning rate for critic & 0.001 \\
        \hline
        $|B|$ & Batch size & 256 \\
        \hline
        $\gamma$ & Discounted factor & 0.99 \\
        \hline
        $\sigma$ & Exploration noise & 0.2 \\
        \hline
        $\eta$& Coefficient of disturbance norm & 10 \\
        \hline
    \end{tabular}
\end{table}
All algorithms were trained using the same set of hyperparameters shown in the table.
For RDDPG, the scalar coefficient $\eta$ multiplying the disturbance norm was fixed to $10$.
For PR-MDP and NR-MDP, the adversarial noise injection probability $\alpha_{\mathrm{AR}}$ was set to $0.1$.
\end{document}